\documentclass[12pt]{article}
\usepackage{setspace}
\usepackage{verbatim}
\usepackage{tabularx}
\usepackage{colortbl}
\usepackage{natbib}
\usepackage{hhline}

\usepackage{multirow}

\usepackage{times} 
\usepackage[top=2.5cm, bottom=2.5cm, left=2.5cm, right=2.5cm]{geometry}
\usepackage{natbib}
\usepackage{amsmath}
\usepackage{amssymb}
\usepackage{latexsym}
\usepackage{sectsty}
\usepackage{amsfonts}
\usepackage{epsfig}
\usepackage{url}
\usepackage{subfigure}
\usepackage{lineno}
\usepackage{wrapfig}

\usepackage{amsmath}
\usepackage{amssymb}
\usepackage{latexsym}
\usepackage{sectsty}
\usepackage{amsfonts}
\usepackage{epsfig}
\usepackage{url}
\usepackage{subfigure}
\newcommand{\zap}[1]{}

\usepackage{color}   

\usepackage{color}   
\newcommand{\sout}[1]{ } 

\title{Soft Triangles for Expert Aggregation}
\author{Paul B. Kantor \\ Paul B. Kantor, Consultant \\ paul.kantor@rutgers.edu }
\date{September 02, 2019}

\begin{document}
\maketitle
\begin{abstract}
We consider the problem of eliciting expert assessments of an uncertain parameter. The context is risk control, where there are, in fact, three uncertain parameters to be estimates. Two of these are probabilities, requiring the that the experts be guided in the concept of ``uncertainty about uncertainty.'' We propose a novel formulation for expert estimates, which relies on the range and the median, rather than the variance and the mean. We discuss the process of elicitation, and provide precise formulas for these new distributions. 
\end{abstract}

\section{Introduction}

There are many important questions for which we lack objective numerical data, and must reply on aggregation of expert opinion to estimate the quantities of interest. Examples include: estimating the number of person who have a particular illness, but have not sought treatment; estimating the number of persons who have attempted some particular crime  and have not been detected; and so forth.  In the arena of Homeland Security, a particularly pressing example of this problem is estimating the risk of attacks, and, perhaps even more pressing, the benefits of various possible policies and procedures to prevent those attacks, or to diminish the harm that they cause. There is a canonical formulation, in the field of risk analysis, which quantifies risk as the product of three ``factors:'' 
\begin{equation}\label{eq:wordRiskequation}
Risk=Threat \times Vulnerability \times Consequences
\end{equation}
Many authors have noted that this formulation admits a translation of Equation  \ref{eq:wordRiskequation}  into the language of utility theory, by viewing $Threat$ as a probability, $Vulnerability$ as a conditional probability, and $Consequences$ as some measure on an appropriate utility scale. The corresponding equation would describe the risk associated with a specific threat $T$ as shown in Equation \ref{eq:risk as probability}.

\begin{equation}\label{eq:risk as probability}
R(t)=U(Success|t)Prob(Success|t;d) Prob(t) 
\end{equation}

Note that Equation \ref{eq:risk as probability} introduces a new and important variable, $d$ which labels the defensive measure(s) that are in place to reduce the risk. 

The key challenge for elicitation is that none of the factors in this equation are concrete physical characteristics that could be known with sufficiently precise measurement. It is (fortunately)  also true  that in the United States we have very little historical experience on the impact of terror-related attacks.  This means that the experts are being asked to estimate something that is fundamentally uncertain. In other words they are being asked to describe a stochastic variable. Everyone who has taught elementary statistics knows that  concept  is not widely understood. This means that effective elicitation must begin with some kind of training activity to introduce the respondents both to the idea of a probability, and to the idea that probabilities are described by giving some kind of a distribution. There are many ways to approach this problem. 

\subsection{Triangular}\label{subs:triangular}

Intuitively, it is fairly attractive to use triangular distributions. The elicitation then goes something like this:

``let's begin by asking what you think is the most likely value of whatever it is we are talking about.''

After some discussion of that one asks:

``and what is the highest it could reasonably be?''

And after that one asks:
``and what is the lowest it could possibly be?''

\begin{figure}
\centering
\includegraphics[width=8.5cm]{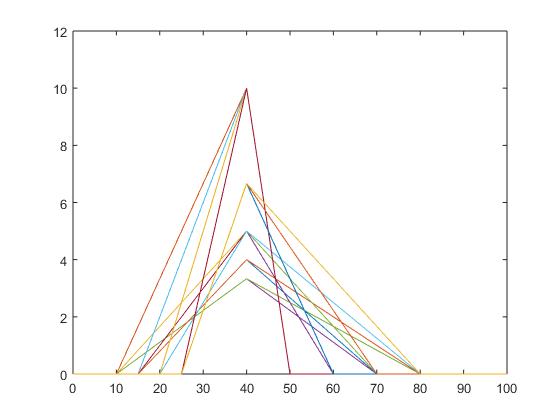}
\caption{Alternate shapes of the `triangle uncertainty distribution. See text.}
\label{fig:triangles}
\end{figure}

These results are then summarized for the respondent by displaying a triangular graph, with the peak at the value that was said to be most probable. Examples of such shapes  are shown in  Figure 
\ref{fig:triangles}. For clarity, all of the distributions have the same mode, while the extremes of the range vary. Unless the distribution is symmetric, the mode is {\bf not} the median or 50-50 point. 

\begin{figure}
\centering
\includegraphics[width=8.5cm]{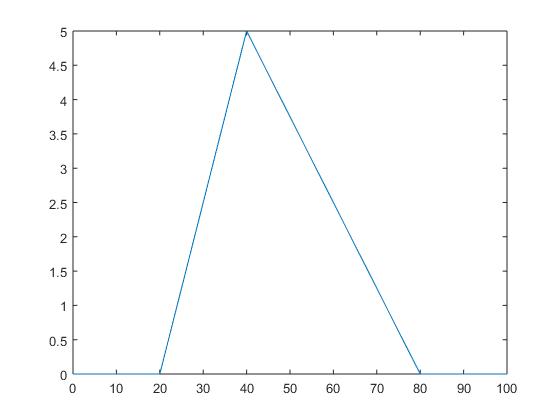}
\caption{An asymmetric triangular distribution. See text.}
\label{fig:20-40-80}
\end{figure}

This  approach can quickly lead to difficulties, when we are interacting with real experts. Let's first consider the triangular distribution.  if an expert has said that the most likely value is 40, and the low extreme is 20, but the high extreme is 80, the resulting distribution look something like Figure \ref{fig:20-40-80}. 

If we regard this as representing the distribution of the unknown variable we note that it is three times as  likely to be above the mode as below. We might try to get around this problem, by actually asking the expert to estimate the median. That could be done with a question something like:

``can you give me a kind of 50-50? That is, the real answer is as likely to be above it is below?"

It would not be surprising, in fact, if the respondent gave exactly the same number that he or she gave when asked for the most likely value. Since people have very little experience estimating probabilities, it would  be surprising if they estimated with the kind of mathematical precision that we attribute to their responses.

\subsection{Beta Distributions}\label{subs:beta}

\begin{figure}
\centering
\includegraphics[width=8.5cm]{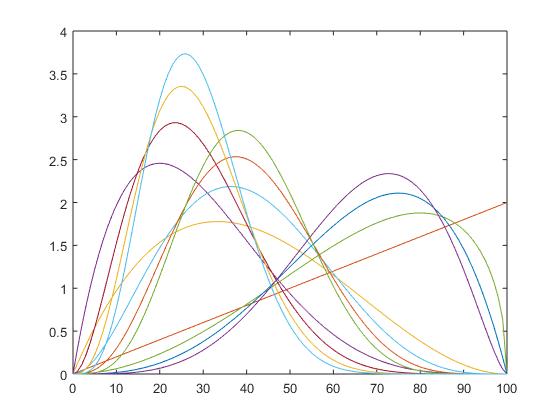}
\caption{Alternate shapes of the Beta  uncertainty distribution; see text.}
\label{fig:someBetas}
\end{figure}

Because much of the work that is done with these probabilities involves Bayesian updating, the triangular distribution is not particularly convenient. With those calculations in mind it is quite natural to prefer the Beta distribution, which updates very gracefully. That is, it updates on the basis of new information. Examples of beta distributions are shown in Figure \ref{fig:someBetas}.

In Figure \ref{fig:someBetas}  $a=1,2,3,4,5$ and $b=.5a,a,3a$. As you can see, these distributions, even when shown graphically, do not capture the intuitive notion of a range (to approximate that one needs high values of $a,b$, which makes it doubtful that they correspond to the subject matter expert's perception of his or her own uncertainty.

Even dealing with the beta distribution, if we actually plot it for the respondents they will see something that looks rather like the statistical distributions they might recall from high school or college, except that  they have neither a lower limit nor an upper limit.

Beta distributions  are of course completely characterized by the mean and the variance. Similarly they can be characterized by the mode and the variance. Elicitation on these models, as developed for example by  \cite{stillwell1987comparing}  typically talks about the mode, or most likely value, and tries to elicit something that serves to pin down the variance. This can  be done by asking a question about the ``width'' of the distribution, which is then translated by a reasonable formula into a statement about the variance, to  finally  pin down the two underlying parameters of the beta distribution. 

Elicitation can also be done using a spreadsheet tool, which displays the distribution, and can be adjusted until the respondent accepts it. We cannot know whether acceptance indicates that the curve is a good match to the mental model of the respondent, or simply means that ``it  is time to get on to the next step of the elicitation.'' \footnote{Of course, the same threat to validity is present for all of the other ways of eliciting probability distributions! }

Now, when dealing with risk assessment, as Equation \ref{eq:risk as probability} shows, we  easily find ourselves with three different uncertain variables that must  be estimated. In this case, and particularly if we plan to  aggregate the opinions of several  experts, it may be useful to have another way of dealing with the elicitation. 

Here are some  desiderata for representing uncertainty:
\begin{enumerate}
   \item It should still make sense to the respondents. Thus, if we can preserve some aspects of the triangular elicitation, that will help.
   \item  It should be tractable when we have to do calculations.
   \item If we are doing small group elicitation, or focus group, it seems helpful  that the elicited distributions, and their aggregation, not impose an unjustified appearance of consensus when such a consensus is lacking.
\end{enumerate}

In experiments with various natural ways of combining estimated distributions (such as weighting  them by confidence and  computing a weighted average) the triangular distribution preserves some signs of the lack of consensus, as it will always be piecewise linear with multiple peaks. The beta distribution does not behave quite so nicely, as things tend to smooth out in complicated ways.

\section{Related Research}\label{sec:lit review}

Elicitation of uncertain parameters, in terms of some distribution, has been considered in several contexts.  Early work was done by \cite{stillwell1987comparing,keeney1991eliciting,kadane1988separating}. For a thorough recent review, see \cite{o2006uncertain}. A non-technical discussion of the importance and difficulty of such elicitation is given by \cite{burgman2016trusting}. 
Finally, a tool that supports group elicitation (although without aggregation) under a number of parametric models is described by \cite{morris2014web}.

Techniques for socializing non-technical experts, such as law enforcement or security personnel have been developed by Stephen Hora of CREATE (personal communication). These begin with some discussion of probability, and its relation to the more familiar concept of ``odds.'' Techniques are often proprietary, and cannot be discussed further here. 

When uncertainties must be combined, as in the conventional formula for risk, Equation \ref{eq:risk as probability}, it may happen that experts place their uncertainty in different parts of the formula, while coming to much the same conclusion. This raises some difficult problems, which we address briefly, in Section \ref{sec:DiscussConclude}.

\section{The Soft Triangle Model}\label{sec:soft triangle}

With these desiderata and difficulties in mind, we propose here a new tool for elicitation that  might be called the ``soft triangle model.''
\subsection{Sharp Soft Triangles}\label{subs:sharp}
In the soft triangle model, we ask for the lower value, the median (or 50-50 point), and the high value. Of course we cannot now simply draw a triangle. 
Instead,  we  first consider  the shorter or narrower side of the distribution, and interpolate the probability linearly on that side. Then, on the wider side, we use a simple monomial interpolation, which falls to zero at the extreme value, and joins smoothly to the linear portion, at the apex. Let us call this distribution $p_{\sharp}$.

The mathematics is not complicated, and the details are given Equations \ref{eq:narrow} {\it et seq.}.

For a monomial, falling to zero at the extreme. There are two parameters. Without loss of generality we will suppose that the long side is the upper part of the ``soft triangle''.  The corresponding monomial is given by 

\begin{align}\label{eq:narrow}
n&=M-L  \\
u(x)&=\frac{H-x}{H-M}  \\
\alpha &=  \\
p_{\sharp}(x) &=nu(x)^\alpha 
\hspace{.5in}   for M \leq x  \leq H
\end{align}

The two conditions, that the curve be continuous, and that the area on both sides of the median, $M$, uniquely determine the exponent $\alpha$.

These concepts have been coded in Matlab, with a script that will also run in the open source version, Octave. For use by the widest community of practitioner experts, it would be desirable to port the code  to Excel, making use of  knobs or sliders. If  multiple experts  participate in  a process   it will also be desirable to make this available as some kind of a Java-based tool, running on a server, or perhaps making use of  a cloud framework such as Google Sheets, although that tool does not, at this time, support knobs or sliders.)

\subsection{Wide Soft Triangles}\label{subs:wide}

While this form is attractive, experimentation shows that it is fairly tight.  If what the expert means by providing the more remote limit is ``it might {\it conceivably} be as far off as this'' then the tightness captures the expert's judgment. If, on the other hand, the limit means something like ``I'm really not that sure, and it might be way over at the limit'' then something much more slowly falling would be appropriate. Since it must fall off, to preserve the median, we can explore a number of slowly falling distributions.  

The following mathematical form is  suggested, primarily because it  look reasonable'' and can be easily coded, without requiring an embedded equation solver.

Making the longer side of the distribution wide is a little more delicate. We propose using a monomial here, as well, but adjusting the value at the extreme to ensure that half of the distribution lies on the long side of the median at $M$. Let us call this function $p_{\flat}$. It has four parameters: $L,M,H,\phi$.

\begin{align}\label{eq:wide}
w&=\max \{M-L,H-M\}  \\
n&=\min \{M-L,H-M\}  \\
x_l&=(x-L)/(M-L) \\
x_r&=(H-x)/(H-M) \\
B&=(1-\phi)/(2w) \\
A&=1/n -B \\ 
u(x)&=\frac{H-x}{H-M}  \\
\alpha &=2(1/n -B)w/\phi  -1 \\  
p_{\flat}(x) &=B+Au(x)^{\alpha}     
\hspace{.5in}   for \hspace{.5in } M \leq x  \leq H
\end{align}

\begin{figure}
\centering
\includegraphics[width=8.5cm]{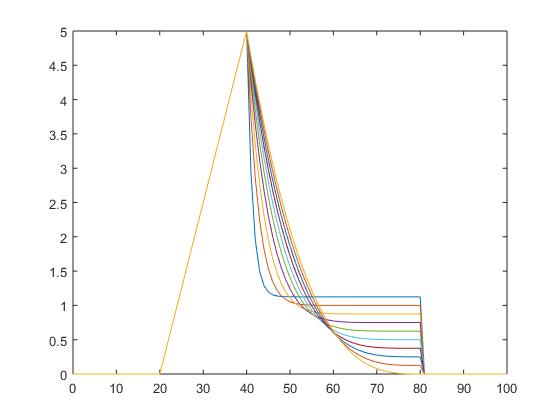}
\caption{Alternate shapes of the ``heavy-tailed'' uncertainty distribution, from $\phi$=0.1 (the broadest) to $\phi=1.0$ (the sharpest). The increment is $\Delta \phi =0.1$. See text.}
\label{fig:wideTails}
\end{figure}

The way these distributions look is shown in Figure \ref{fig:wideTails}. As discussed in Section \ref{sec:eliciting}, the experts will be shown the way that this distribution responds to adjustment of the $\phi$ parameter, and asked to make a choice. Those who are comfortable with a very sharp ($\phi=1$) formulation will have a somewhat simpler task. 

\section{Eliciting Soft Triangle Distributions}\label{sec:eliciting}

After a customary discussion about the way that we estimate probabilities, and a few example cases (such as estimating the number of street intersections an nearby major city; or estimating how many people in your state have library cards; or what the population of some foreign country is) we ask each participant to estimate the low, median (or 50-50 point) and high parameters for the distribution. We then display both the wide and narrow distributions for some typical set of values, and ask each participant separately to indicate whether he favors the narrow or wide version. Elicitation text might be: ``if you set that extreme just to cover all the bases, then narrow may be better; if you really think that this value could just as well be at the extreme as at the 50-50 then you want the wide version.'' 

Those who choose the narrow version can take a break. Elicitation would continue, for the others, as: ``If you chose the wide version we will give you a little more control, to bring more of the distribution or chance near the middle. We do that with a parameter we call ``phi.'' If phi is very small it means you are going wide. As phi move up towards 1, you are pushing all of the chances towards the 50-50 point.'' 

At this point, on the first and second elicitation, we show some examples, or a graph such as Figure \ref{fig:wideTails}. In effect, each expert who has chose the wide option is given a  ``slider'' to adjust the sharpness. 

\subsection{Alternative Models}

Even within this space of half-linear, monomial models, there is a complementary class of models, in which the wider side of the distribution is made linear, while the narrow side is monomial. In that case, $\alpha$ would  always be less than one (to get more area in less space). This amounts to concentrating  the uncertainty around the median itself. while this might be attractive to some participants, it tends to destroy the nice ``each expert has his own peak in the graph'' property, and for this reason we do not explore it further.  

\section{Calculating with soft triangle distributions}\label{sec:Calculating}
Typical calculations in the field of risk-elicitation, and assessment of countermeasures  involve computing the product of several random factors. A good example is risk calculations of the form: 

\begin{align}\label{eq:risk}
\mathbf{R=CVT}    
\end{align}

where each of the threat (T), vulnerability (V) and consequences (C) are uncertain, and should be treated as distributions. In some special cases the distribution of a product of independent random variates can be expressed in terms of known special functions, and a few cases in which the variates are not independent have been recently found in more complex formulation, typically as infinite sums. A summary is given in \cite{wikipedia_product_2020}.

For the distributions proposed here, it is far simpler to do the computations numerically. The necessary relation is:

\begin{align}\label{eq:productDistribution}
\Pr\{XY \le t\}=  \int_0^{\infty}{ dx \int_0^{t/x} Pr\{X=x   \; \& \; Y=y\}dy}
\end{align}

For the probabilities of threat, and the conditional probabilities of success given the threat, it is not hard to do the double integral numerically (the upper limit is not $\infty$, but $1$), for a set of discrete values of $t \in [0,1]$. The corresponding density function is then found by numerical differentiation, for clarity in presentation. 

\section{Discussion and Conclusion}\label{sec:DiscussConclude}

Elicitation in unfamiliar territory carries many risks. As the old joke has it, if we ask 1,000 people to estimate the size of the emperor's nose, the standard error of the mean might be quite small, even if the population know nothing about the emperor. However, any information at all (in the joke, the size of the emperor's shoes) will lend the estimate {\it some} credibility. In the field of risk assessment, the most confident estimators seem to lie \footnote{For insurers, errors on the high side in risk estimation carry no corporate risk, until some competitors are able to reason more accurately and lower their premiums.} in the insurance industry, perhaps in more than one sense of the word. 

We are more interested in optimizing the allocation of scarce resources to defensive measures, and therefore need estimates that are either accurate, or all inflated by the same factor. The comparison of allocations is driven by the ratios of Risk to Cost, so that a numeraire makes no difference, whether it is applied to the Cost, to the Risk, or to both.  

\begin{figure}
\centering
\includegraphics[width=8.5cm]{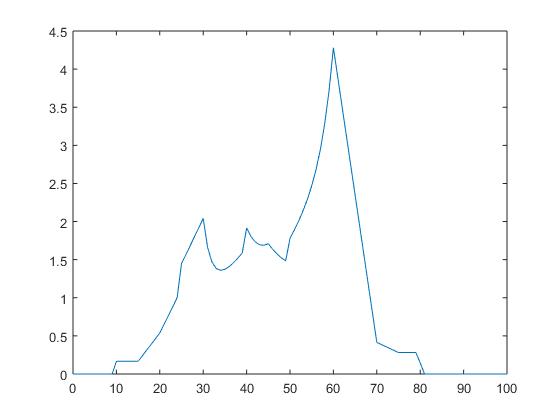}
\caption{Combined distribution for six notional experts. The abscissa is the probability to be estimated (as a percentage); the ordinate is the combined probability distribution itself,
The plot does have some peaks, but cannot resolve the two experts who both put the median at 60\%. 
 }
\label{fig:6experts}
\end{figure}

We give an  example of what the combined plot (using unweighted averages) looks like in the case of 6 experts with a wide range of views. Their data are given in Table \ref{tab:6experts}, and the resulting graph is shown in Figure \ref{fig:6experts}.

\begin{table}[h!]
\begin{center}
 \begin{tabular}{|| r | r| r|r| r|| } 
 \hline
 Expert &Low  & Median  & High  & Phi \\ [0.5ex] 
 \hline\hline
1 & 20	&	40	&	80	&	0.3	\\
2 & 50	&	60	&	70	&	1	\\
3 & 10	&	45	&	70	&	0.3	\\
4 & 15	&	30	&	79	&	0.2	\\
5 & 25	&	50	&	75	&	0.01	\\
6 & 40	&	60	&	70	&	0.4	\\
 [1ex] 
 \hline
\end{tabular}
\caption{Data for six notional subject matter experts. }
\label{tab:6experts}
\end{center}
\end{table}

Finally, we note that if the goal is compare countermeasures, and we are looking for the best among several choices, we can finesse the problem of determining the magnitude of the risk, and ask experts directly: ``how much do you believe that each specific countermeasure reduces the risk.'' Unfortunately, we must usually consider more than one risk, and when the risks are to be summed, they must all have a common numeraire, with regard to both costs and impacts.

As with all approaches to a complex problem, the methods proposed here will, no doubt, require minor, and perhaps major adjustments when they encounter the behavior of actual subject matter experts.

\section{Acknowledgments}\label{sec:acknowledgments}

The author thanks many colleagues at CCICADA, and at CREATE for thoughtful discussions of risk and its estimation.Particular thanks to Vladimir Menkov for correcting an error in the first version of this paper. Thanks also to the Department of Industrial and Systems Engineering at the University of Wisconsin, for support and a courtesy appointment.

\bibliographystyle{plainnat}
\bibliography{ThisBib}

\begin{thebibliography}{7}
\providecommand{\natexlab}[1]{#1}
\providecommand{\url}[1]{\texttt{#1}}
\expandafter\ifx\csname urlstyle\endcsname\relax
  \providecommand{\doi}[1]{doi: #1}\else
  \providecommand{\doi}{doi: \begingroup \urlstyle{rm}\Url}\fi

\bibitem[Burgman(2016)]{burgman2016trusting}
Mark~A Burgman.
\newblock \emph{Trusting judgements: how to get the best out of experts}.
\newblock Cambridge University Press, 2016.

\bibitem[Kadane and Winkler(1988)]{kadane1988separating}
Joseph~B Kadane and Robert~L Winkler.
\newblock Separating probability elicitation from utilities.
\newblock \emph{Journal of the American Statistical Association}, 83\penalty0
  (402):\penalty0 357--363, 1988.

\bibitem[Keeney and Von~Winterfeldt(1991)]{keeney1991eliciting}
Ralph~L Keeney and Detlof Von~Winterfeldt.
\newblock Eliciting probabilities from experts in complex technical problems.
\newblock \emph{IEEE Transactions on engineering management}, 38\penalty0
  (3):\penalty0 191--201, 1991.

\bibitem[Morris et~al.(2014)Morris, Oakley, and Crowe]{morris2014web}
David~E Morris, Jeremy~E Oakley, and John~A Crowe.
\newblock A web-based tool for eliciting probability distributions from
  experts.
\newblock \emph{Environmental Modelling \& Software}, 52:\penalty0 1--4, 2014.

\bibitem[O'Hagan et~al.(2006)O'Hagan, Buck, Daneshkhah, Eiser, Garthwaite,
  Jenkinson, Oakley, and Rakow]{o2006uncertain}
Anthony O'Hagan, Caitlin~E Buck, Alireza Daneshkhah, J~Richard Eiser, Paul~H
  Garthwaite, David~J Jenkinson, Jeremy~E Oakley, and Tim Rakow.
\newblock \emph{Uncertain judgements: eliciting experts' probabilities}.
\newblock John Wiley \& Sons, 2006.

\bibitem[Stillwell et~al.(1987)Stillwell, Von~Winterfeldt, and
  John]{stillwell1987comparing}
William~G Stillwell, Detlof Von~Winterfeldt, and Richard~S John.
\newblock Comparing hierarchical and nonhierarchical weighting methods for
  eliciting multiattribute value models.
\newblock \emph{Management Science}, 33\penalty0 (4):\penalty0 442--450, 1987.

\bibitem[Wikipedia(2020)]{wikipedia_product_2020}
Wikipedia.
\newblock Product distribution, July 2020.
\newblock URL
  \url{https://en.wikipedia.org/w/index.php?title=Product_distribution&oldid=970273508}.
\newblock Page Version ID: 970273508.

\end{thebibliography}

\end{document}